\documentclass[letterpaper, 10 pt, conference]{ieeeconf}  

\IEEEoverridecommandlockouts                           

\usepackage{graphicx} 
\usepackage{amsmath} 
\usepackage{multirow}

\title{\LARGE \bf
Deep Attention Driven Reinforcement Learning (DAD-RL) for Autonomous Decision-Making in Dynamic Environment
}

\author{Jayabrata Chowdhury$^{1*}$, Venkataramanan Shivaraman$^{2*}$, Sumit Dangi $^{3*}$, Suresh Sundaram $^{2}$, P B Sujit $^{4}$
\thanks{This work was supported by Qualcomm Innovation Fellowship India 2023. The authors marked with * contributed equally.}
\thanks{$^{1}$Jayabrata Chowdhury is with Robert Bosch Centre for Cyber-Physical Systems, Indian Institute of Science, Bangalore.
        {\tt\small jayabratac@iisc.ac.in}}%
\thanks{$^{2}$Venkataramanan Shivaraman and $^{2}$Suresh Sundaram are with the Department of Aerospace Engineering, Indian Institute of Science, Bangalore.
        {\tt\small vshivaraman18@gmail.com, vssuresh@iisc.ac.in}}%
\thanks{$^{3}$Sumit Dangi and $^{4}$P.B. Sujit are with the Department of Data Science \& Engineering and Electrical Engineering \& Computer Science, Indian Institute of Science Education and Research, Bhopal.
        {\tt\small sumit19@iiserb.ac.in, sujit@iiserb.ac.in}}%
}

\begin{document}
\maketitle
\thispagestyle{empty}
\pagestyle{empty}

\begin{abstract}
Autonomous Vehicle (AV) decision-making in urban environments is inherently challenging due to the dynamic interactions with surrounding vehicles. For safe planning, AV/ego must understand the weightage of various spatiotemporal interactions in a scene. Contemporary works use colossal transformer architectures to encode interactions mainly for trajectory prediction, resulting in increased computational complexity.  To address this issue without compromising spatiotemporal understanding and performance, we propose the simple Deep Attention Driven Reinforcement Learning (DAD-RL) framework, which dynamically assigns and incorporates the significance of surrounding vehicles into the ego’s RL-driven decision-making process. We introduce an AV-centric spatio-temporal attention encoding (STAE) mechanism for learning the dynamic interactions with different surrounding vehicles. To understand map and route context, we employ a context encoder to extract features from context maps. The spatio-temporal representations combined with contextual encoding provide a comprehensive state representation. The resulting model is trained using the Soft-Actor Critic (SAC) algorithm. We evaluate the proposed framework on the SMARTS urban benchmarking scenarios without traffic signals to demonstrate that DAD-RL outperforms recent state-of-the-art methods. Furthermore, an ablation study underscores the importance of the context-encoder and spatio-temporal attention encoder in achieving superior performance.
\end{abstract}

\section{INTRODUCTION}
\begin{figure}[t]
    \centering
    \includegraphics[scale=0.15]{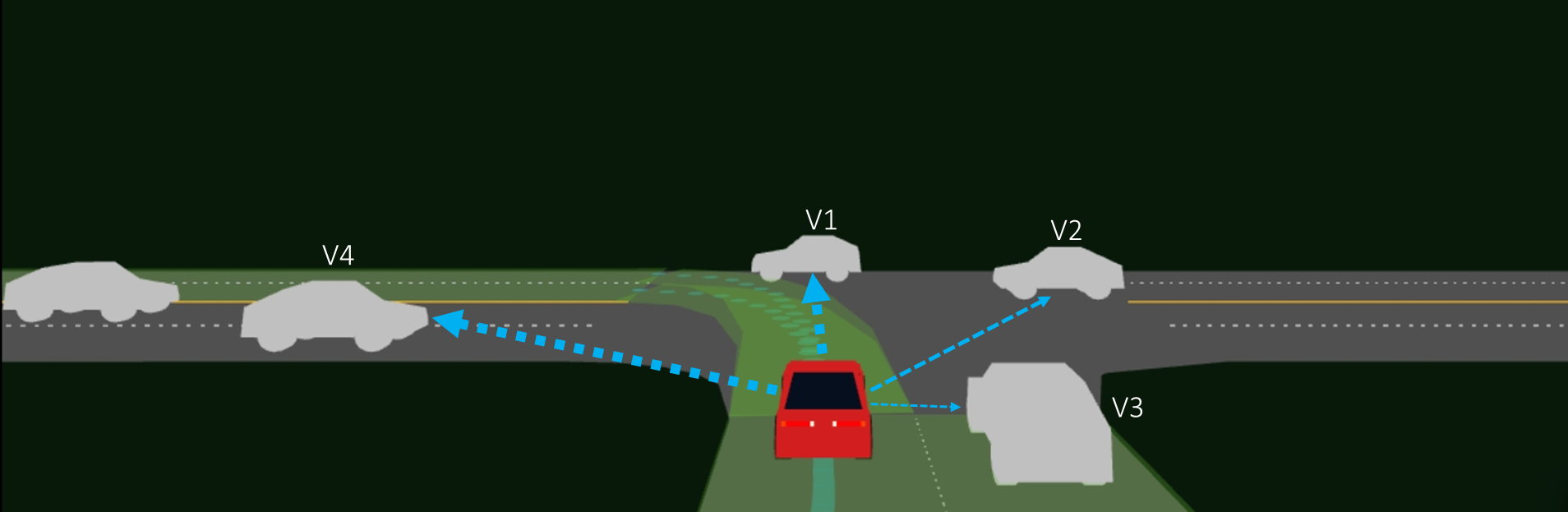}
    \caption{A left-turn scenario with surrounding vehicles. The AV is depicted in red, while the other vehicles are illustrated in white. The desired route is designated in green. The AV must comprehend the significance of each neighboring vehicle in relation to its final objective of reaching the destination, as indicated by blue arrows of varying weights.}
    \label{left_turn}
\end{figure}

Navigating safely in a dynamic environment populated with other vehicles remains a significant hurdle for Autonomous Vehicles (AVs). Decisions made by the AV should not only be safe but also comply with human driving behavior. Fig. \ref{left_turn} illustrates a left-turn scenario without a traffic signal, emphasizing the AV's need to comprehend other road users' actions and decide the attention importance to ensure safe navigation. Previous methods have explored rule-based methods \cite{Aksjonov_2021_Rule_based}. Rule-based methods excel in scenarios the rules were defined for but falter in new ones. An alternative approach \cite{Bazzi_2021_Comm, Cui_2022_Coopernaut} involves explicit communication between the AV and other vehicles, enabling the AV to make informed decisions in collaboration with other vehicles. However, this method is limited because reliable communication channels can only be assured between vehicles from the same manufacturer. For effective decision-making, the AV must comprehend the dynamic driving context as it evolves implicitly. In a dynamic driving environment, AV should understand the temporal behaviors of other surrounding vehicles and learn to make safe decisions. Also, these behaviors can be influenced by spatial structures such as road geometry. 

Imitation Learning approaches involve learning from an expert's actions. Several recent studies (\cite{Chen_2019_DIL, Chowdhury_2022_DST-CAN, Jamgochian_2023_SHAIL, Cai_2021_DiGNet}) have utilized Imitation Learning (IL) methods to develop decision-making abilities that mimic an expert driver. However, expert bias and distribution shift challenges can considerably affect the efficacy of these IL-based approaches. Given the absence of near-collision situations in expert driving, it is difficult for IL-based techniques to recover from such scenarios. Recent advances in Reinforcement Learning (RL) based decision-making algorithms \cite{Kiran_2022_DRL_survey} show promising performance. RL's strength stems from its exploration capabilities; it can recover from near-collision scenarios better. However, these methods need an understanding of the state representations for RL. To better understand the socially related spatio-temporal interactive behaviors between AV and other vehicles, a method must be developed to encode spatio-temporal relationships and provide information for safe decision-making. 

In the context of an AV navigating a roadway, the safety relevance of other vehicles varies. The work in \cite{Zhang_2020_Uncertainty} has identified essential vehicles using rule-based expertise. However, such expert knowledge can be intricate and may only sometimes scale to unfamiliar driving situations. In a dynamic real-world setting, the significance of each vehicle in the vicinity of the AV fluctuates with each passing moment. Therefore, a spatio-temporal state space representation is needed to encapsulate the evolving importance of the interactions between the AV and surrounding vehicles. This work introduces the Deep Attention Driven Reinforcement Learning (DAD-RL) framework to encode the spatio-temporal interactions between the AV and nearby vehicles with contextual information. This framework extracts efficient state-space representation for secure decision-making in a dynamic environment. The temporal encoders encode temporal relationships, aiding in understanding the spatial dynamics of the AV and each surrounding vehicle. To encode socially interactive behaviors, we utilize the ego AV-oriented attention mechanism \cite{Vaswani_2017_Attention}. This mechanism is instrumental in learning crucial spatio-temporal features for decision-making based on RL.

As proposed, the DAD-RL framework offers a solution to model the dynamic spatio-temporal interaction between AV and its surrounding vehicles for decision-making. The key contributions are:
\begin{enumerate}
    \item DAD-RL introduces an innovative approach to model the ego AV-centric attention mechanism. The query vector associated with AV dynamically learns the attention to surrounding vehicles' key and value vectors through a spatio-temporal attention encoder. 
    \item A context encoder is developed to extract the contextual features important for AV. The final state encoder combines both encodings. This serves as the state space representation for RL-based decision-making.
    \item A dense reward structure is designed to help in safe and efficient decision-making.
    \item The DAD-RL surpasses the performance of the recent larger transformer-based model, Scene-Rep-T (SRT) \cite{Liu_2024_Scene-Rep-T}, in SMARTS \cite{Zhou_2020_SMARTS} in terms of success rate, collision, and stagnation. An overall improvement of $29.6 \%$ and $2.4 \%$ in success rate compared to SAC and SRT, respectively. Furthermore, an ablation study on decisions with errors in human-like behavior and overall score underscores the significance of context encoding with the spatio-temporal context attention mechanism.
\end{enumerate}
\begin{figure*}[t]
    \centering
    \includegraphics[scale=0.32]{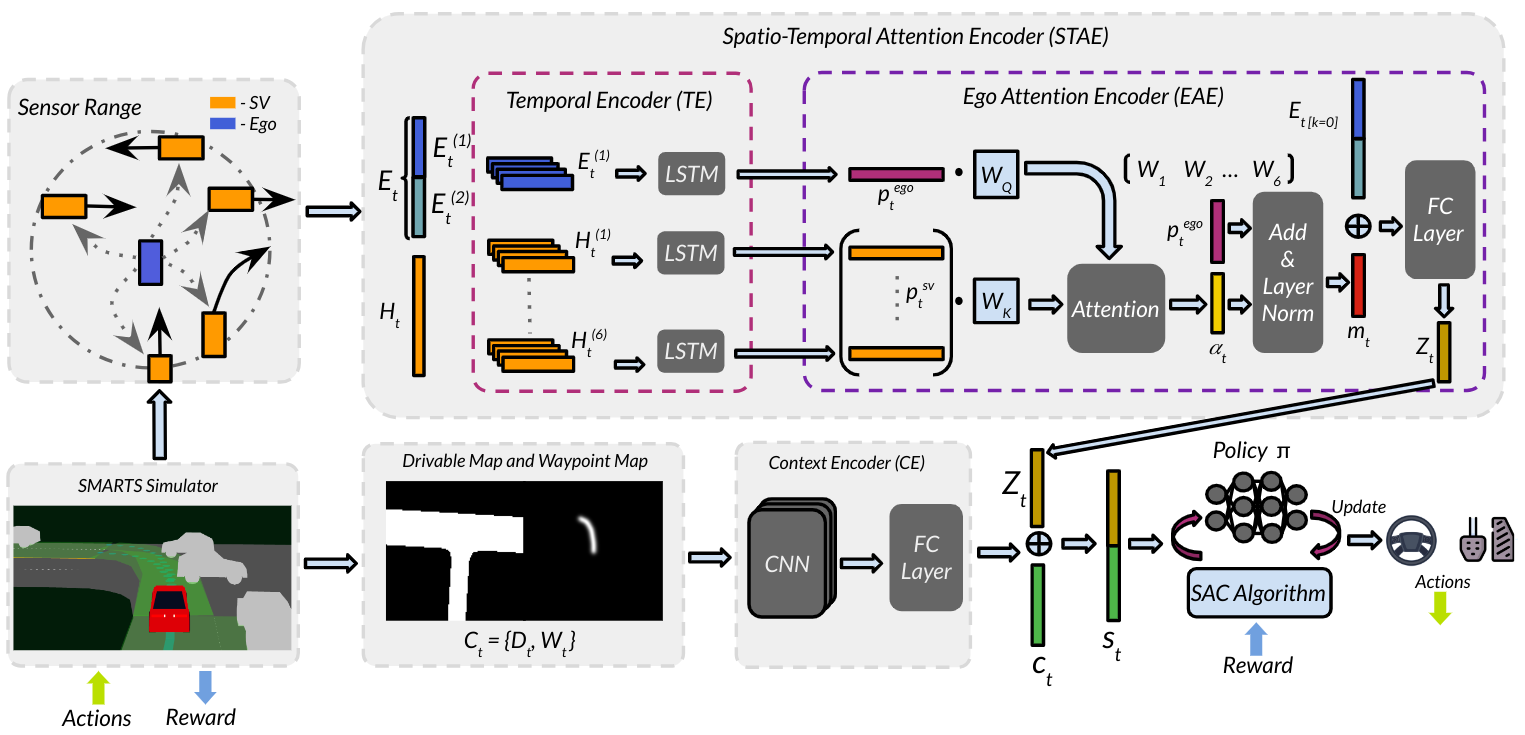}
    \caption{The schematic diagram showing the components of the DAD-RL framework. It graphically shows the observation space from the SMARTS simulator, the Spatio-Temporal Attention Encoder, the Bird-Eye-View context encoder, and the action space $V^{target}_t$ and $\Lambda_t$.}
    \label{fig: Schematic_diagram_of_DAD_RL}
\end{figure*}

\section{RELATED WORK}
In the contemporary research landscape, the autonomous decision-making capabilities of Reinforcement Learning (RL) have been utilized for autonomous driving applications \cite{Chen_2022_Interpretable_DRL, Tang_2022_Highway, Zhang_2023_Multi}. However, these studies necessitate a spatio-temporal state representation that can understand the dynamics between the Autonomous Vehicle (AV) and its surrounding vehicles. Recent research \cite{Janer_2021_Offline_RL, Chen_2021_Decision_Transformer} has utilized a Transformer encoder-decoder architecture to model the sequential decision-making process, effectively transforming RL decision-making into conditional sequence modeling. Recognizing the importance of interactive behavior modeling for social navigation, certain studies \cite{Schmidt_2022_Crystal_graph, Wang_2024_Graph_traj} have employed Graph Neural Networks (GNNs) for predicting trajectories. Similarly, acknowledging the significance of spatio-temporal interaction modeling for decision-making, some research has incorporated graph-based models \cite{Cai_2022_DQ-GAT, Chowdhury_2024_GP3Net} for state encoding in RL-based decision-making. Our work utilizes a streamlined and modified self-attention mechanism to model the state space representation between the AV and its surrounding vehicles. It is subsequently used for RL-based decision-making.

\section{THE DAD-RL FRAMEWORK}
This section explains the mechanics of the DAD-RL decision-making framework. The primary component of DAD-RL is the spatio-temporal deep attention state-encoding mechanism learned and wielded by RL-driven ego-vehicle. The framework also consists of a context-encoder to process route information. This RL driving task is formulated as a POMDP since our ego-vehicle can access limited knowledge ascribed to sensor range limitations. The following subsections explain the input and observation space design, the DAD-RL processing, and the RL framework with action space and reward structure. 

\subsection{Observation Space (Input) Preprocessing}
The ego-vehicle can obtain historical information of its surrounding vehicles $H_{t}$, segmented Bird-Eye-View (BEV) context $C_{t}$, and an elaborate ego's odometric state history $E_t$ as shown in Fig. \ref{fig: Schematic_diagram_of_DAD_RL}. The whole observation space tensor is represented as $\mathcal{O}_t = \left[H_t;E_t;\mathcal{C}_t\right]$. The subscript $t$ attributes to the tensor's value at time $t$. The BEV context is defined as $\mathcal{C}_{t_{o}} = \{D_t; W_t\}$, where $D_t$ is drivable area map and $W_t$ waypoint map of size $128 \times 128$. The historical surrounding vehicle data is defined as ${H}_{t} = \left[H^{1}_{t}, ..., H^{n}_{t} \right]$ for $n$ surrounding vehicles, with $H^{i_{sv}}_{t} = \left [\left ( X^{i_{sv}}_{t - 5k.\delta t}, \phi^{i_{sv}}_{t - 5k.\delta t}, v^{i_{sv}}_{t - 5k.\delta t}, l^{i_{sv}}_{t - 5k.\delta t} \right) \right]^{4}_{k=0}$ of vehicle $i$'s current and past timesteps spaced four simulation steps $\delta t$ apart. Here, for vehicle $i$, $X^{i_{sv}}_{t} = (x^{i_{sv}}_{t}, y^{i_{sv}}_{t})$, $\phi^{i_{sv}}_{t}$, $v^{i_{sv}}_{t}$, and $l^{i_{sv}}_{t}$ are historical data of the relative position to the ego, the heading, speed of the vehicle and the lane it is in respectively. Observation $E_t = \left [ \left ( E^{(1)}_{t}; E^{(2)}_{t} \right) \right]$ consists of various historical odometric values split into two vectors based on utility in the DAD-RL framework. The vector $E^{(1)}_{t}$ contains the same quantities as in $H^{i_{sv}}_{t}$ and $E^{(2)}_{t} = \left [ \left ( \omega_{t - 5k.\delta t}, \psi_{t-5k.\delta t}, v^{ego}_{t - 5k.\delta t}, a^{ego}_{t - 5k.\delta t}, j_{t - 5k.\delta t} \right ) \right ]^{4}_{k =0}$, a tuple having steering, yaw rate, linear velocity, linear acceleration, and linear jerk of the ego at time $t$ respectively. The following section explains how the spatiotemporal deep attention encoder processes this observation space. The encoder has two parts: a) Spatio-Temporal Attention Encoder (STAE) and b) Context Encoder (CE). 

\textbf{Spatio-Temporal Attention Encoder (STAE):} This encoder $\rho_\eta$ takes in the historical kinematic states of the ego and surrounding vehicles $E_t$ and $H_t$ at timestep $t$ as input. To encode the temporal kinematic relationships (past 2.1 seconds or 21 $\delta t$ simulation steps) of a dynamic traffic scenario, the states of the surrounding vehicles $H^{i_{sv}}_{t}$ and the ego vehicle $E^{(1)}_{t}$ is passed through a shared Long Short-Term Memory (LSTM) network. Let's denote the aforementioned vectors of each vehicle $H^{i_{sv}}_{t}$ and ego $E^{(1)}_{t}$ as $I_{t}$. Eq \ref{LSTM_equation} represents the intermediate temporal encoding $p_{t}$. 
\begin{equation}
    \begin{aligned}
        I^{i}_t = E^{(1)}_t \mbox{or} \  H^{i_{sv}}_t \\
        p^{i}_t = LSTM(I^{i}_{t})
    \end{aligned}
     \label{LSTM_equation}
\end{equation}
The temporal encoding for ego $p^{ego}_{t}$ and $p^{i_{sv}}_{t}$ for all $I_{t}$ can be found from the hidden states of the LSTM. The ego vehicle's attention has been modeled based on the temporal kinematic encodings. This has been obtained using an attention mechanism between each temporal encoding. Let $W_{q}$, $W_{k}$, and $W_{v}$ are the query, key, and value weight matrices, respectively. To employ the attention mechanism, $p^{ego}_t$ is considered as the query $Q$ and $p^{i_{sv}}_t$ for all $n$ surrounding vehicles as both the key $K$ and the value $V$ vectors since it is important for the ego encoding $p^{ego}_t$ to pay attention to each surrounding vehicle encoding $p^{i_{sv}}_t$. The following Equ.\ref{attention_equ} shows the employed attention mechanism. Let query, key and value be $ Q = p^{ego}_t; K = V = p^{1:n_{sv}}_t $ and $\alpha_t$ be attention.
\begin{equation}
    \begin{aligned}
     \alpha_t(Q, K, V) = \sigma \left (\frac{Q.W_q[K.W_k]^T + M}{\sqrt{d_k}} \right ) V
    \end{aligned}
    \label{attention_equ}
\end{equation}
The softmax or the $\sigma(Q, K)$ term outputs the attention weights for each vehicle to every other vehicle present. However, since the interest is on the attention weights associated with the ego on other vehicles, only the first row of the matrix in the $\sigma(Q, K)$ is taken. After the attention block, the final spatio-temporal attention $Z_t$ is derived as given in the Equ.\ref{state_encoding}, where $\it{Norm}$, $\oplus$, and $\it{Linear}$ represent layer normalization, concatenation method, and fully connected layer, respectively. $E_{t[k=0]}$ is ego state vector at current time instant. $M$ is a mask that hides absent vehicles from the attention calculation. This mask helps in cases where the number of surrounding vehicles in the sensor range is less than the maximum number of vehicles (n).
\begin{equation}
    \begin{aligned}
        m_t = Norm \ (\alpha_t + p^{ego}_{t}) \\
        {Z_ t} = Linear \ (E_{t[k=0]} \oplus m_{t})
    \end{aligned}
    \label{state_encoding}
\end{equation}

\textbf{Context Encoder (CE):} To process the BEV context maps $\mathcal{C}_t$, which contain drivable area map $D_t$ and waypoint map $W_t$, which are images, a Convolutional Neural Network (CNN) is used. To obtain a concise vector, CNN layers are employed in a manner that decreases the image size after each successive layer. After the CNN layers, the intermediate output is flattened and passed through a fully connected layer, giving the context encoding vector $c_t$. After obtaining spatio-temporal attention encoding $Z_t$ of the surrounding vehicles and odometric states of the ego and the context encoding $c_t$, the vectors $Z_t$ and $c_t$ are concatenated to obtain a final state encoding $s_t$ as shown in Equ.\ref{final_encoder}. 
\begin{equation}
    s_t = Z_t \oplus c_t
    \label{final_encoder}
\end{equation}

\subsection{RL Algorithm}
The ego vehicle has a stochastic policy network $\pi_{\theta}$ with $\theta$ as its parameters mapping state $s_t$ to actions. As in any RL task, the objective is to learn $\pi_{\theta}$ along with our STAE ($\rho_\eta$) that can maximize the cumulative reward obtained by the ego-vehicle. Soft Actor-Critic (SAC), a state-of-the-art RL algorithm, is used for training policy networks and STAE. 

\textbf{Reward Structure:} The ego-vehicle is trained using a dense reward structure for safety and comfort (evaluated by ‘Humanness error’ metrics). The reward structure (Equation \ref{Reward_Structure}) is a linear combination of rewards and penalties:
\begin{equation}
    \begin{aligned}
            R = \lambda_1.r_{crash} + \lambda_2.r_{road} + \lambda_3.r_{v^{ego}} + \lambda_4.r_{goal} \\ + \lambda_5.r_{prog}  + \lambda_6.r_{oroute} + \lambda_7.r_{ww} + \lambda_8.r_{slow}  
    \end{aligned}
    \label{Reward_Structure}
\end{equation}
The positive reward terms are $r_{goal}$ and $r_{prog}$. $r_{goal} = 1$ if the agent reaches the goal; otherwise, $r_{goal} = 0$; $r_{prog}$represents distance travelled towards goal. To encourage the ego vehicle's momentum towards reaching the goal under speed limits, the reward is defined as $r_{v^{ego}} = v^{ego}/V_{max}$ when $v^{ego} < V_{max}$ and a penalty defined as $r_{v^{ego}} = -\mbox{abs}(v^{ego} - V_{max})/V_{max}$ when the ego vehicle is overspeeding. The penalty terms are defined as $r_{crash} = r_{road} = r_{route} = r_{ww} = r_{slow} = -1$, representing penalty for crashing, going offroad, off route and wrong way, and for not moving for ten consecutive seconds. The $\lambda_{i}$ coefficients are used to scale up/down and adjust the weights of each term in the reward structure.

\subsection{Action Space (Output) Representation}
The action space is mid-level control, combining continuous and discrete actions. Mathematically, the action space is defined as $\mathcal{A}_t = \left [V_t^{target}, \Lambda_t \right ]$, where $V_t^{target} \in (0, V_{max})$ represents the target speed of the vehicle at time $t$ and $\Lambda_t$ represents lane commands such as 'switch to left/right lane' or 'keep lane.' A classical controller present in the SMARTS simulator executes these middle-level commands. The SAC algorithm's policy network $\pi_\theta$ is designed for continuous action spaces. To accommodate discrete lane change commands, $\pi_\theta$ which outputs parameters of a Gaussian Distribution from which lane actions are sampled in the range $(-1,1)$, is divided into partitions a) $(-1, -1/3)$, b) $(-1/3, 1/3)$ and c) $(1/3, 1)$, where a) and c) are mapped to 'switch to left/right lane' respectively and b) to 'keep lane'. 

\section{EXPERIMENTS and RESULTS}
\subsection{Driving Scenarios}
The DAD-RL framework strongly emphasizes utilizing interaction encoding techniques to comprehend intricate and realistic traffic settings. SMARTS has been chosen as the simulation platform to assess the effectiveness of the DAD-RL framework in handling such complex scenarios. Within SMARTS, several demanding scenarios were constructed aimed at both training and testing the DAD-RL approach. These scenarios are designed to encompass diverse interactive and stochastic traffic dynamics. The following scenarios capture various aspects of realistic driving behaviors.


\textbf{Left Turn-T:} This is an urban T-junction with heavy traffic and no traffic signals. The goal of the ego-vehicle is to take an unprotected left turn.


\textbf{Roundabout:} In this urban roundabout scenario, the ego transitions from a 2-lane bi-directional road to a 2-lane unidirectional roundabout with four exits—three versions: Roundabout-A, B, and C, with increasing difficulties in that order. The ego vehicle is supposed to cover a quarter, half, and three-quarters in Roundabout-A, B, and C, respectively. The commute distance sets the difficulty level.

\textbf{Double-Merge:} In this scenario, an autonomous vehicle ('ego') starts from a single-lane road, navigates a two-lane one-way road with two entrances/exits, and exits on the opposite side. The 'ego' must effectively perform lane changes amidst traffic flows from all entrances to exits, honing its navigation and lane-changing skills.

The scenarios above incorporate heavy traffic flows randomly selected for each simulation episode. The agent engages in multiple consecutive scenarios to assess its overall performance during evaluation.

\subsection{Training Setup}
This subsection explains how the DAD-RL framework is trained. The SMARTS simulator is used as an RL-Gym environment for training. Each simulation step is $\delta t = 0.1s$ apart. The ego-vehicle obtains information about surrounding vehicles, itself, and the context at every step. The ego-vehicle processes the observation, takes action, and proceeds to the next simulation step. A route is initialized for the vehicle to follow at the beginning of each episode. Each step consists of a tuple $(\mathcal{O}_t, a_t, \mathcal{O}_{(t+1)}, r_t, d_t)$ with $d_t$ being done/terminal signal. A series of such steps and tuples make one episode. The DAD-RL framework is trained on five scenarios mentioned in the previous subsection. A vectorized environment runs several simulations parallelly to speed up the training process. As several steps/episodes progress, the ego-vehicle stores the experience tuple $(\mathcal{O}_{0:B}, a_{0:B}, \mathcal{O}_{0:B}, r_{0:B}, d_{0:B})$ in a buffer B. After gaining sufficient experience, a batch is sampled from this buffer B to estimate the objective function, and the loss is back-propagated from the end of the policy network $\pi_\theta$ till the beginning of STAE $\rho_\eta$ based on loss functions of SAC. After this network training process, the ego-vehicle collects more experience, repeating the process. The training was done on a machine with at least 32 GB RAM, an Nvidia 2080Ti GPU, and Ubuntu 20.04. 

\subsection{Comparison Baselines}
To thoroughly compare the proposed framework's performance, the DAD-RL framework was evaluated against several baseline methods, and SOTA \cite{Liu_2024_Scene-Rep-T}. \textbf{PPO}: Proximal Policy Optimization stands out as a SOTA policy gradient approach with promising performance on robotic decision-making tasks. \textbf{SAC}: Soft Actor-Critic is another SOTA RL algorithm, an off-policy approach capitalizing on entropy maximization to assist agent training. \textbf{DrQ}: Data-regularized Q-learning employs a CNN-based augmentation to SAC, facilitating direct learning of a policy function from pixels. \textbf{RDM}: Rule-based Driver Model mimicking the driving behavior of neighboring vehicles within SMARTS simulations, offering a benchmark for evaluating the rule-based performance. \textbf{DT}: Decision Transformer is an innovative approach that reformulates the RL problem as a conditional sequence modeling task and employs a transformer to generate future actions from past states, actions, and rewards. \textbf{SRT}: Scene-Rep Transformer (SRT) is a transformer-based framework recently proposed in \cite{Liu_2024_Scene-Rep-T}. SRT employs two transformer architectures - a Multi-Stage Transformer (MST) for encoding the multi-modal scene input and another Sequential Latent Transformer (SRT) to instill predictive information into the latent state vector.

For all the mentioned baselines, the implementation provided by \cite{Liu_2024_Scene-Rep-T} is considered for the performance comparison, and the results are directly sourced from the same. 

\subsection{Comparative Results}
In evaluating the proposed framework, several performance metrics were utilized to measure the experiments' effectiveness. These metrics provide insights into the agent's behavior and performance. Here is an explanation of the performance metrics adopted for the evaluation: 
\textbf{Succ.\% (Success Rate)}: This metric represents the proportion of episodes where the agent successfully reaches the goal out of the total evaluation episodes. A higher success rate indicates better performance in goal achievement. \textbf{Coll.\% (Collision Rate)}: The collision rate is the proportion of episodes that resulted in a collision between the agent and surrounding vehicles. A lower collision rate signifies better navigation and avoidance capabilities. \textbf{Stag.\% (Stagnation Rate)}: Stagnation rate reflects the proportion of episodes ending prematurely due to exceeding the maximum time steps allowed. A lower stagnation rate indicates efficient decision-making and goal pursuit.
\textbf{Humanness Error}: Humanness Error is calculated based on comfort factors such as jerk and angular acceleration. It assesses how closely the agent's movements resemble human-like driving behavior. \textbf{Overall Score}: Overall Score depicts a comprehensive measure of driving performance. As implemented in SMARTS, it combines multiple factors such as progress, rule violations, and comfort.
Humanness Error and Overall Score are both provided by SMARTS. To assess the performance of the DAD-RL framework based on the defined metrics, we follow an evaluation strategy similar to \cite{Liu_2024_Scene-Rep-T}. The evaluation strategy involves playing 50 episodes in simulation to test the agent's performance across varied traffic patterns, enabling a comprehensive analysis of the proposed framework's effectiveness. The initialization for the testing environment is randomly selected.

Table \ref{tab:results} shows the results of three evaluation metrics, namely success rate, collision rate, and stagnation rate for all the baselines mentioned in the previous section and the DAD-RL framework. It is evident from the results that the DAD-RL framework exhibits significant performance improvements compared to the baseline methods in most of the scenarios. DAD-RL shows exceptional performance on the left turn scenario, with a $0\%$ collision rate. DAD-RL significantly improves the success and collision rates on all roundabout scenarios compared to the SRT. The trend of success rates in the three roundabout scenarios reflects their difficulties. As we go from R-A to R-B to R-C, the route length increases, which means the ego will face more traffic interactions, increasing the difficulty.
In the Double Merge scenario, DAD-RL faces tough competition from MST and SRT, yet it still surpasses other baselines like DT and DrQ in terms of overall performance.
Unlike the other four scenarios, Double Merge requires the agent to assess the perfect time to merge into a lane with ongoing traffic to avoid collisions. This could explain the discrepancies in results for Double Merge, and it can be solved with a spatiotemporal prediction module similar to SRT.
On average, across all five scenarios, DAD-RL increases the success rate by $\textbf{29.6\%}$ compared to SAC and $\textbf{2.4\%}$ compared to SRT. These quantitative findings emphasize the superior capabilities of DAD-RL in completing driving tasks and mitigating collision incidents. 
The architectural difference between DAD-RL and SRT is noteworthy, with the latter leveraging a full Transformers model while the former opts for a simpler approach using a compact spatio-temporal encoder with a single attention layer. This streamlined model design enhances efficiency by concentrating on crucial learning components. The lightweight nature of the DAD-RL framework not only delivers enhanced average performance and simplifies architectural complexity, making it a practical and effective choice for driving tasks.

\begin{table}
\caption{Comparative Results On Different Scenarios}
\label{tab:results}
\centering
\begin{tabular}{|c|c|ccc|}
\hline
\multirow{2}{*}{Scenario}     & \multirow{2}{*}{Model} & \multicolumn{3}{c|}{Metric}                                           \\ \cline{3-5} 
                              &                        & \multicolumn{1}{c|}{Succ.\% $\uparrow$} & \multicolumn{1}{c|}{Coll.\% $\downarrow$} & Stag.\% $\downarrow$ \\ \hline
\multirow{8}{*}{Left Turn-T}  & RDM                    & \multicolumn{1}{c|}{2}       & \multicolumn{1}{c|}{54}      & 44      \\ \cline{2-5} 
                              & PPO                    & \multicolumn{1}{c|}{36}      & \multicolumn{1}{c|}{50}      & 10      \\ \cline{2-5} 
                              & SAC                    & \multicolumn{1}{c|}{68}      & \multicolumn{1}{c|}{28}      & 0       \\ \cline{2-5} 
                              & DrQ                    & \multicolumn{1}{c|}{78}      & \multicolumn{1}{c|}{20}      & 0       \\ \cline{2-5} 
                              & DT                     & \multicolumn{1}{c|}{66}      & \multicolumn{1}{c|}{32}      & 0       \\ \cline{2-5} 
                              & MST                    & \multicolumn{1}{c|}{88}      & \multicolumn{1}{c|}{12}      & 0       \\ \cline{2-5} 
                              & SLT                    & \multicolumn{1}{c|}{94}      & \multicolumn{1}{c|}{4}       & 0       \\ \cline{2-5} 
                              & \textbf{DAD-RL}                 & \multicolumn{1}{c|}{\textbf{98}}      & \multicolumn{1}{c|}{\textbf{0}}       & \textbf{0}       \\ \hline
\multirow{8}{*}{Roundabout-A} & RDM                    & \multicolumn{1}{c|}{68}      & \multicolumn{1}{c|}{30}      & 0       \\ \cline{2-5} 
                              & PPO                    & \multicolumn{1}{c|}{66}      & \multicolumn{1}{c|}{34}      & 0       \\ \cline{2-5} 
                              & SAC                    & \multicolumn{1}{c|}{76}      & \multicolumn{1}{c|}{24}      & 0       \\ \cline{2-5} 
                              & DrQ                    & \multicolumn{1}{c|}{80}      & \multicolumn{1}{c|}{20}      & 0       \\ \cline{2-5} 
                              & DT                     & \multicolumn{1}{c|}{76}      & \multicolumn{1}{c|}{22}      & 0       \\ \cline{2-5} 
                              & MST                    & \multicolumn{1}{c|}{84}      & \multicolumn{1}{c|}{16}      & 0       \\ \cline{2-5} 
                              & SLT                    & \multicolumn{1}{c|}{88}      & \multicolumn{1}{c|}{12}      & 0       \\ \cline{2-5} 
                              & \textbf{DAD-RL}                 & \multicolumn{1}{c|}{\textbf{96}}      & \multicolumn{1}{c|}{\textbf{4}}       & \textbf{0}       \\ \hline
\multirow{8}{*}{Roundabout-B} & RDM                    & \multicolumn{1}{c|}{2}       & \multicolumn{1}{c|}{98}      & 0       \\ \cline{2-5} 
                              & PPO                    & \multicolumn{1}{c|}{42}      & \multicolumn{1}{c|}{58}      & 0       \\ \cline{2-5} 
                              & SAC                    & \multicolumn{1}{c|}{48}      & \multicolumn{1}{c|}{52}      & 0       \\ \cline{2-5} 
                              & DrQ                    & \multicolumn{1}{c|}{72}      & \multicolumn{1}{c|}{28}      & 0       \\ \cline{2-5} 
                              & DT                     & \multicolumn{1}{c|}{68}      & \multicolumn{1}{c|}{32}      & 0       \\ \cline{2-5} 
                              & MST                    & \multicolumn{1}{c|}{76}      & \multicolumn{1}{c|}{24}      & 0       \\ \cline{2-5} 
                              & SLT                    & \multicolumn{1}{c|}{82}      & \multicolumn{1}{c|}{18}      & 0       \\ \cline{2-5} 
                              & \textbf{DAD-RL}                 & \multicolumn{1}{c|}{\textbf{88}}      & \multicolumn{1}{c|}{\textbf{12}}      & \textbf{0}       \\ \hline
\multirow{8}{*}{Roundabout-C} & RDM                    & \multicolumn{1}{c|}{0}       & \multicolumn{1}{c|}{100}     & 0       \\ \cline{2-5} 
                              & PPO                    & \multicolumn{1}{c|}{38}      & \multicolumn{1}{c|}{50}      & 12      \\ \cline{2-5} 
                              & SAC                    & \multicolumn{1}{c|}{46}      & \multicolumn{1}{c|}{48}      & 6       \\ \cline{2-5} 
                              & DrQ                    & \multicolumn{1}{c|}{68}      & \multicolumn{1}{c|}{30}      & 2       \\ \cline{2-5} 
                              & DT                     & \multicolumn{1}{c|}{66}      & \multicolumn{1}{c|}{30}      & 0       \\ \cline{2-5} 
                              & MST                    & \multicolumn{1}{c|}{66}      & \multicolumn{1}{c|}{34}      & 0       \\ \cline{2-5} 
                              & SLT                    & \multicolumn{1}{c|}{76}      & \multicolumn{1}{c|}{24}      & 0       \\ \cline{2-5} 
                              & \textbf{DAD-RL}                 & \multicolumn{1}{c|}{\textbf{80}}      & \multicolumn{1}{c|}{\textbf{20}}      & \textbf{0}       \\ \hline
\multirow{8}{*}{Double Merge} & RDM                    & \multicolumn{1}{c|}{0}       & \multicolumn{1}{c|}{100}     & 0       \\ \cline{2-5} 
                              & PPO                    & \multicolumn{1}{c|}{36}      & \multicolumn{1}{c|}{64}      & 0       \\ \cline{2-5} 
                              & SAC                    & \multicolumn{1}{c|}{62}      & \multicolumn{1}{c|}{22}      & 0       \\ \cline{2-5} 
                              & DrQ                    & \multicolumn{1}{c|}{76}      & \multicolumn{1}{c|}{14}      & 0       \\ \cline{2-5} 
                              & DT                     & \multicolumn{1}{c|}{70}      & \multicolumn{1}{c|}{30}      & 0       \\ \cline{2-5} 
                              & MST                    & \multicolumn{1}{c|}{92}      & \multicolumn{1}{c|}{4}       & 0       \\ \cline{2-5} 
                              & SLT                    & \multicolumn{1}{c|}{\textbf{96}}      & \multicolumn{1}{c|}{\textbf{2}}       & 0       \\ \cline{2-5} 
                              & \textbf{DAD-RL}                 & \multicolumn{1}{c|}{86}      & \multicolumn{1}{c|}{14}      & \textbf{0}       \\ \hline
\end{tabular}
\end{table}

\subsection{Ablation Study}
\begin{figure}[t]
    \centering
    \begin{tabular}{@{}c@{}}
         \includegraphics[scale=0.14]{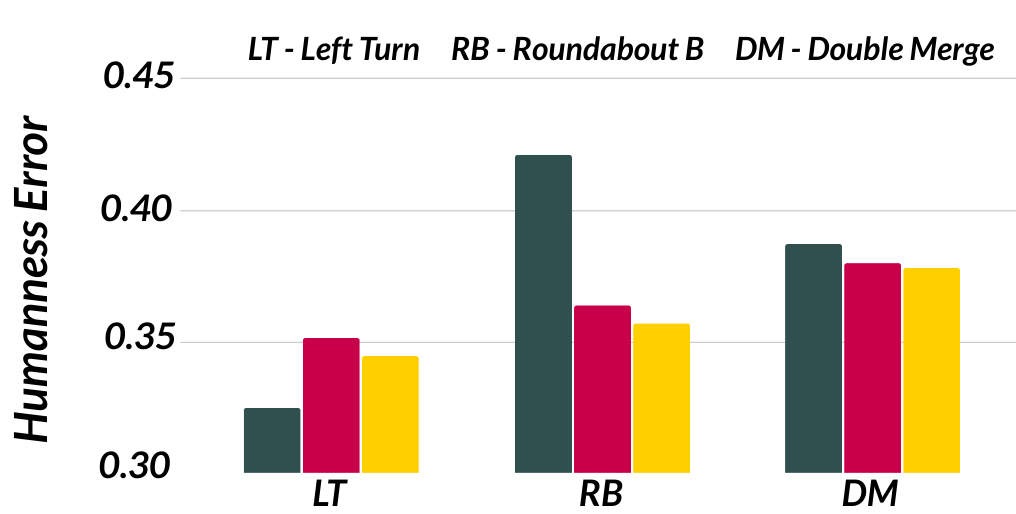}
    \end{tabular}
    \begin{tabular}{@{}c@{}}
        \includegraphics[scale=0.20]{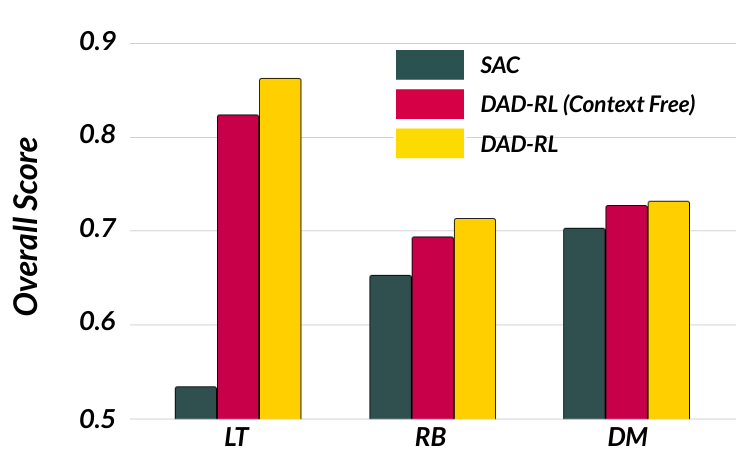}
    \end{tabular}
    \caption{Plots for (a) Humanness Error and (b) Overall Score for different scenarios. Context-free is DAD-RL without a context encoder module.}
    \label{fig:Overall}
\end{figure}

Multiple experiments on three distinct scenarios were conducted to understand the individual aspects of the spatio-temporal encoder and the context encoder. The results of these experiments are depicted in Fig. \ref{fig:Overall}.
Both the Roundabout and Double Merge scenarios exhibit similar trends among the metrics. Compared to SAC, which only utilizes a context encoder, using only the proposed spatio-temporal encoder (Context) improves the overall score while also reducing humanness errors. A combination of the spatio-temporal attention encoder and the context encoder (DAD-RL) achieves the highest Overall Score and the lowest humanness error. 
In the Left Turn-T scenario case, the trends are predominantly consistent with the other scenarios except for one metric. SAC obtains a lower humanness error than both variants of DAD-RL.
Upon visualizing SAC's performance in simulation, it was noticed that the agent shows conservative behavior, making it difficult for the agent to maneuver through the traffic at the T-junction, leading to a lower Overall score. This conservative driving style is the reason for the lower humanness error of SAC. DAD-RL shows a slight compromise regarding humanness error to improve the overall score greatly.

\section{CONCLUSIONS}
This paper underscores the challenges recent approaches and RL-based decision-making algorithms encounter in dynamic driving environments. While emphasizing the significance of interaction modeling, we introduce a simple spatio-temporal attention encoder rather than a full transformer for secure decision-making in AV. A new approach to attention modeling for AVs is introduced, which encodes dynamic interactions with surrounding vehicles using the Deep Attention Driven Reinforcement Learning (DAD-RL) framework. The framework enhances performance compared to previous state-of-the-art RL-based decision-making for AVs, including the work that uses a transformer. Future research will concentrate on integrating the dynamics of AVs with safety layer design and interpretable decision-making, paving the way for more robust and reliable autonomous systems.

\addtolength{\textheight}{-12cm}   



\end{document}